\documentclass[letterpaper]{article} 
\usepackage{aaai25}  
\usepackage{times}  
\usepackage{helvet}  
\usepackage{courier}  
\usepackage[hyphens]{url}  
\usepackage{graphicx} 
\usepackage{natbib}  
\usepackage{caption} 
\usepackage{algorithm}
\usepackage{algorithmic}
\usepackage{multirow}
\usepackage{amsmath,amsbsy}
\usepackage{amsfonts}
\usepackage[toc,page,header]{appendix}

\usepackage{newfloat}
\usepackage{listings}
\DeclareCaptionStyle{ruled}{labelfont=normalfont,labelsep=colon,strut=off} 
\floatstyle{ruled}
\newfloat{listing}{tb}{lst}{}
\floatname{listing}{Listing}
\title{iMoT: Inertial Motion Transformer for Inertial Navigation}
\author{
    Son Minh Nguyen\textsuperscript{\rm 1},
    Linh Duy Tran\textsuperscript{\rm 2},
    Duc Viet Le\textsuperscript{\rm 1},
    Paul J.M Havinga\textsuperscript{\rm 1}
}
\affiliations{

    \textsuperscript{\rm 1}Department of Computer Science, University of Twente, Enschede, The Netherlands \\
    \textsuperscript{\rm 2}Viettel AI, Viettel Group, Hanoi, Vietnam\\
     \textsuperscript{\rm 1}\{m.s.nguyen; v.d.le; p.j.m.havinga\}@utwente.nl;\,\,\textsuperscript{\rm 2}linhtd15@viettel.com.vn
}

\usepackage{bibentry}

\begin{document}

\maketitle

\begin{abstract}

 We propose iMoT, an innovative  Transformer-based inertial odometry method that retrieves cross-modal information from motion and rotation modalities for accurate positional estimation. Unlike prior work, during the encoding of the motion context, we introduce Progressive Series Decoupler at the beginning of each encoder layer to stand out critical motion events inherent in acceleration and angular velocity signals. 
To better aggregate cross-modal interactions, we present Adaptive Positional Encoding, which dynamically modifies positional embeddings for temporal discrepancies between different modalities.
During decoding, we introduce a small set of learnable query motion particles as priors to model motion uncertainties within velocity segments. Each query motion particle is intended to draw cross-modal features dedicated to a specific motion mode, all taken together allowing the model to refine its understanding of motion dynamics effectively. Lastly, we design a dynamic scoring mechanism to stabilize iMoT's optimization by considering all aligned motion particles at the final decoding step, ensuring robust and accurate velocity segment estimation. Extensive evaluations on various inertial datasets demonstrate that iMoT significantly outperforms state-of-the-art methods in delivering superior robustness and accuracy in trajectory reconstruction.
\begin{links}
  \link{Code}{https://github.com/Minh-Son-Nguyen/iMoT}
\end{links}

 

  


\end{abstract}

\section{Introduction}

Inertial odometry systems are intended to estimate the three-dimensional trajectories of a moving body by jointly analyzing motion, rotation, and geographically magnetic interactions from Inertial Measurement Unit (IMU) signals. Given privileges typified by energy efficiency, privacy preservation, and environmental robustness over other modalities, these approaches become more essential for tracking and tracing missions, including virtual and augmented reality, biodiversity monitoring, and rescue operations, especially in harsh environments, such as areas shrouded in smoke, or mist where common modalities (e.g., radio, acoustic, and visual signals) are no longer reliable for tracking purposes. 

Contemporary approaches to inertial navigation are commonly classified into three distinct research paradigms: physics-based methods, heuristic priors-based methods, and data-driven priors-based methods. Since quadratic errors are leaked through the double integral of noisy IMU measurements whilst computing displacements, classical physics-based methods,  exemplified by strap-down inertial navigation systems (SINS) \cite{titterton2004strapdown} result in huge drift accumulation in a very short time.
Heuristic priors-based methods, represented by pedestrian dead reckoning (PDR) approaches \cite{jimenez2009comparison,tian2015enhanced}, decompose trajectory estimation into discrete components: step detection, step-length estimation, and heading estimation, all 
 operating under assumptions of regular human gait patterns. However, these methods often face limitations in adaptability, with step detection and step-length estimation being confined to specific scenarios and occasionally relying on fixed parameters. Moreover, heading estimation may suffer from inaccuracies induced by gravitational and magnetic disturbances. In an advanced manner, data-driven priors-based methods \cite{chen2018ionet,herath2020ronin,liu2020tlio} utilize end-to-end deep learning architectures to directly estimate velocity segments from IMU sequences, significantly reducing drifting errors in trajectory reconstruction. Despite their progress, these methods fail to consider modality distinctions between acceleration and angular rates, as well as motion uncertainties among individuals, resulting in compromised accuracy. 

In this paper, we present a multimodal \textbf{i}nertial \textbf{Mo}tion \textbf{T}ransformer (iMoT), which maximizes the utilization of complementary features between motion and rotation inputs to dynamically represent motion uncertainties for different instantaneous velocity segments. 
Firstly, we address the practical challenges of input tokenization in iMoT. 
Conventional tokens captured at a single time step with multiple variates (i.e., channels) often struggle to convey a complete chain of motion events due to excessively local receptive fields and time-unaligned events represented by concurrent time points \cite{zhang2023crossformer}. 
In contrast, we consider the entire time series of each variate as an input token.

Secondly, given the multivariate time series nature of IMU sequences, we extend the time series decomposition \cite{hyndman2018forecasting}, a widely adopted pre-processing technique, into Progressive Series Decoupler (PSD), a trainable module that can be seamlessly integrated within encoder layers. Specifically, PSD enhances unimodal features of acceleration and angular velocity signals and aids the absorption of such information by progressively decomposing their intricate temporal patterns into more interpretable components, which better highlight critical motion events such as half-turns, U-turns, and periods of stillness.
 To reflect modality differences, we propose Adaptive Positioning Encoding (APE), which associates acceleration and angular tokens with appropriate positional embeddings based on their temporal contents. These fully embedded tokens are then processed through a self-attention module to capture multivariate correlations. Additionally, we introduce Adaptive Spatial Sync (ASC) at every residual connection to retain fine-grained details across channels.

Thirdly, we recognize that motion modes can vary between individuals and over time, affecting both the direction and magnitude of instantaneous velocity segments. Then, each velocity segment can be defined by the most likely combination of these motion modes. Inspired by Particle Filter techniques, we model motion uncertainties for a given velocity segment by manipulating a set of learnable query motion particles during decoding.
Specifically, each query motion particle represents the velocity of a specific motion mode, functioning as a learnable positional embedding to probe related cross-modal features. This probing initiates a cyclic process, where the cross-modal features extracted in each decoding step are continuously used to inform and refine the query particles in successive steps. Finally, we propose a Dynamic Scoring Mechanism (DSM) to optimize the query particle set during both training and testing phases.

Our main contributions are four-fold: (1) We propose both novel inertial Transformer encoder and decoder networks (iMoT) that harness the complementary effects of motion and rotation modalities for uncertainty modeling to enhance positional accuracy. (2) In encoding context features, we first introduce Progressive Series Decoupler (PSD) to highlight motion events within each modality, followed by Adaptive Positional Encoding (APE) to include distinctions between modalities in position encoding. Additionally, we implement Adaptive Spatial Sync (ASC) at every residual connection to ensure the seamless integration of spatial details. (3) In decoding, we introduce a novel concept of query motion particles that utilize cross-modal information retrieval from context features to learn all possible motion modes accounting for uncertainties in motion. At the final decoding step, these adjusted particles are collectively considered to determine the desired velocity segments through a unique Dynamic Scoring Mechanism (DSM). (4) We rigorously validate the contribution of each proposed module and demonstrate the overall effectiveness of iMoT against state-of-the-art (SoTA) odometry methods over four benchmark inertial datasets.

\section{Related Work}
\textbf{Physics-based methods (no priors)}. 
A strap-down inertial navigation system (SINS)\cite{savage1998strapdown} first rotates acceleration measurements from the body frame to the navigation frame using a rotation matrix derived from integrating angular velocity measurements in order to subtract the Earth's gravity. Locations are then acquired by double-integrating the linear acceleration\cite{shen2018closing}. Due to noisy sensing compounded by multiple integrations, these strap-downs incur quadratic error propagation heavily, drifting far away from desired locations. 

\noindent\textbf{Heuristic priors.} To alleviate drifting errors accrued in SINS, step-based pedestrian dead reckoning (PDR) approaches \cite{jimenez2009comparison,tian2015enhanced} leverage human motion regularities by separately detecting steps, estimating step length and heading before updating the location with each step. These approaches yield impressive results under controlled environments where the assumption remains valid.
Some other methods  \cite{janardhanan2014attitude, kourogi2014method} that incorporate principal component analysis and frequency domain analysis have been developed to enhance rotational accuracy by determining body motion directions. However, these heuristic-based methods fall short of robustness, as their inner components are interdependent on gravitational and magnetic factors, and confined to specific scenarios.

\noindent\textbf{Data-Driven priors.}
Significant efforts have recently been directed toward developing deep learning networks to extract useful features from IMU measurements for enhanced position estimation. RIDI \cite{yan2018ridi} introduces a two-stage system that first regresses low-frequency corrections to the acceleration before double-integrating it into displacement. To bypass the noisy double integration, IoNeT \cite{chen2018ionet} employs LSTMs to directly regress polar velocity segments and changes in heading rates, which are then cumulatively computed to approximate translations. Similarly, RoNIN framework \cite{herath2020ronin} explores three types of neural networks to assess their distinct contributions to inertial navigation. Building on RoNIN, TLIO \cite{liu2020tlio} integrates displacement estimates with raw IMU measurements using a stochastic-cloning Extended Kalman Filter (EKF) to derive position, orientation, and sensor biases. More recently, CTIN \cite{rao2022ctin} adapts ResNet blocks into a Transformer architecture, which incorporates motion uncertainties by optimizing covariance of velocity segments.

\noindent\textbf{Transformer.}
Transformers have found extensive applications in natural language processing \cite{vaswani2017attention,devlin2018bert} and computer vision \cite{dosovitskiy2020image,carion2020end}. Our approach is partially inspired by DETR \cite{carion2020end} that utilizes query object tokens during decoding to extract object-related information for detection. Based on this idea, we propose a novel concept of learnable query motion pairs, yet consisting of query motion particles and their associated content features, for modeling motion uncertainties. In an advanced manner, these query pairs, especially the query motion particle set, are not only externally updated in the form of learnable positional embeddings but also undergo internal motion refinement within transformer decoder layers. This dual updating process enables more precise modeling of motion dynamics.


\section{Proposed Method}

\begin{figure*}[t!]
\vspace{-4pt}
\centering
\includegraphics[width=0.74\textwidth] {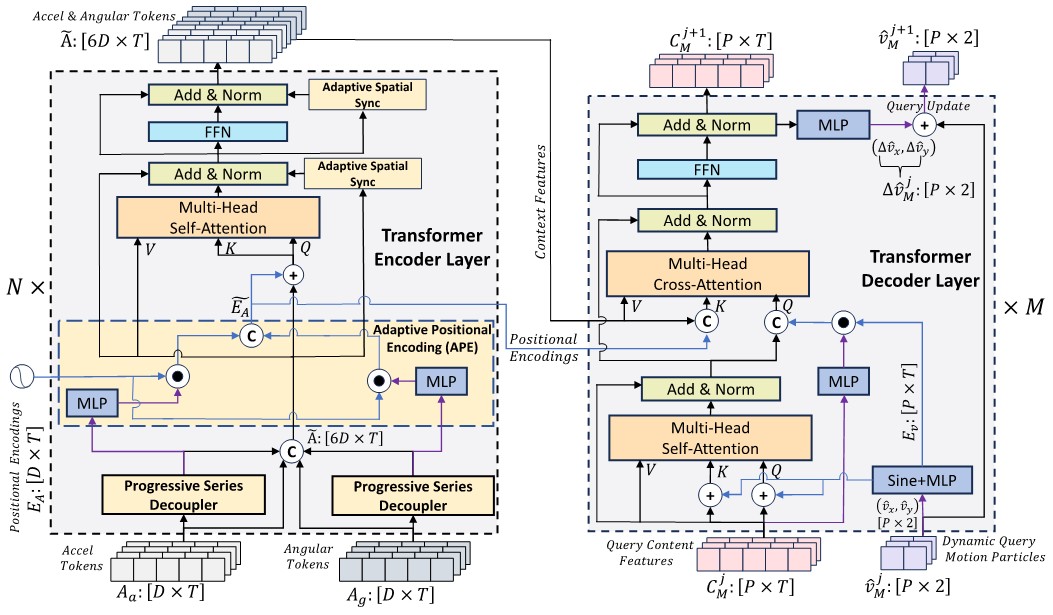} 
\vspace{-10pt}
\caption{
iMoT architecture: (1) The Encoder synthesizes motion context features from Acceleration and Angular Velocity tokens, incorporating three key innovations: Progressive Series Decoupler for enhanced information absorption, Adaptive Positional Encoding to handle modality differences, and Adaptive Spatial Sync to maintain cross-channel interactions. (2) The Decoder manipulates query motion particles to capture motion variability within velocity segments through cross-modal information retrieval. Two signal types are highlighted: the \textit{blue flow} for base signals like sinusoidal encoding and the \textit{magenta flow} for controlling signals, including tokens and velocity particles, making static modules or operations responsive to changes.
}
\label{fig:imot}
\vspace{-15pt}
\end{figure*}

Inertial odometry methods aim to reconstruct a traveled trajectory from corresponding IMU sequences, denoted as $A = \{A_a, A_g\} \forall A \in {\mathbb{R}^{2D \times T}}$, where ${A_a} \in {\mathbb{R}^{D \times T}}$ and ${{\rm{A}}_{\rm{g}}} \in {\mathbb{R}^{D \times T}}$ represent acceleration and angular velocity of $D \times T$ instances recorded over $D=3$ channels along $x$-, $y$-, $z$-axes within $1$ second, respectively. In this work, we propose \textbf{i}nertial \textbf{Mo}tion \textbf{T}ransformer (iMoT), a novel transformer encoder-decoder structure designed to iteratively refine query motion particles to learn all possible motion modes for every single instantaneous velocity segment. This approach allows iMoT to effectively adapt to variations in velocity segments, thereby enhancing trajectory synthesis. The overall structure is illustrated in Fig.\ref{fig:imot}. 
\vspace{-8pt}
\subsection{Encoder}
\vspace{-4pt}
 In practice, all movements are inherently reflected by fluctuations in the motion and rotation modalities within IMU sequences. To seamlessly encode such motion context, we directly reshape raw IMU sequences into input tokens compatible with the transformer encoder. However, existing spatial tokens formed by a single time step $t \in T$ along $D$ channels often struggle to express contextual information due to excessively local receptive fields and time-unaligned events.
In this way, series variations are significantly influenced by the sequence order, which might undermine the attention mechanism when improperly applied to the temporal dimension. As a result, the proposed model may be compromised in its ability to capture essential series representations and portray multivariate correlations, limiting its capacity and generalizability on diverse IMU data.
Therefore, we first tokenize the sequence into temporal variate tokens $A_a$ and $A_g$ of $T$-dimensions, which offer a broad receptive field to capture important events along the temporal axis. Following this, we introduce a series of specialized modules incorporated in encoder layers to enhance the exploitation of these tokens.
\subsubsection{Progressive Series Decoupler}
\begin{figure}[t!]
\vspace{-4pt}
\centering
\includegraphics[width=0.35\textwidth] {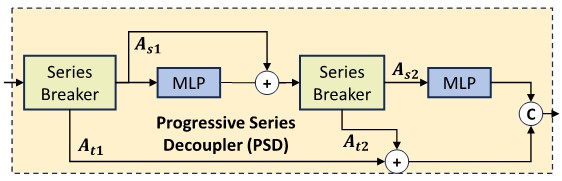} 
\vspace{-10pt}
\caption{Progressive Series Decoupler.}
\label{fig:PSD}
\vspace{-15pt}
\end{figure}
To promote the capture of necessary motion events during encoding, Progressive Series Decoupler (PSD) is introduced at the facade of each encoder layer, breaking down complex temporal patterns of motion and rotation series into two more interpretable components, namely seasonal signals and trend-cycle signals. That is, seasonal signals carry prominent information about symmetric human motions (e.g., leg movements, hip swiveling, and arm swinging), which is particularly useful for tracing repeated sequence patterns (e.g., walking, strolling, and running).  On the other hand, trend-cycle signals exhibit longer-term movements, revealing sudden motion changes (e.g., sudden stopping and turning).
Unlike the prior work \cite{wu2021autoformer}, we design a centered moving average to smooth out periodic fluctuations and eliminate some of the randomness in the data, leaving a smooth trend-cycle component. The seasonal signal is then computed as the residual:
\begin{equation}
\begin{array}{l}
{A_t} = AvgPoo{l_{k_2 \times k_1}}\left( {Padding\left( A \right)} \right)\\
{A_s} = A - {A_t}
\end{array}
\end{equation}
\noindent Here, $A_s \in \mathbb{R}^{2D \times T}$, and $A_t \in \mathbb{R}^{2D \times T}$ denote the seasonal and the extracted trend-cycle parts, respectively. The $Padding$ operation is used to keep the series length unchanged. The $AvgPool_{k_2 \times k_1}(\cdot)$ operation depicts the application of a moving average of order $k_1$ followed by another moving average of order $k_2$. Note that $k_2$ must be less than $k_1$, and both should be either odd or even numbers to ensure the symmetry of the weighted average applied to the observations from both the inner and outer sides.
As depicted in Fig.\ref{fig:PSD}, we denote $A_s, A_t$ = $SeriesBreaker(A)$, an internal operation embedded within PSD module that is progressively learned over layers. After undergoing PSD, sequence tokens $A$ are associated with their decoupled components $A_t$ and $A_s$ together as standard input tokens $\Tilde A \in \mathbb{R}^{6D \times T}$ to the subsequent blocks. This approach helps isolate and interpret different motion cues with much greater ease, enhancing the model's ability to handle complex IMU data effectively.

\subsubsection{Adaptive Positional Encoding} Positional Embeddings play a crucial role in extracting, storing, and pooling context features in both encoding and decoding phases. As presented in Fig.\ref{fig:imot}, we initialize Adaptive Positional Encoding (APE) with common base sinusoidal positional embeddings $E_A \in \mathbb{R}^{D \times T}$. From these base embeddings, we start to learn temporal scaling factors according to content discrepancies between modalities, resulting in adaptive positional embeddings $\widetilde {{E_A}} \in \mathbb{R}^{6D \times T}$:
\vspace{-4pt}
\begin{equation}
\scalebox{0.8}{$
\widetilde {{E_A}} = \left[ {{\rm{MLP}}\left( {\Tilde {{A_a}}} \right) \cdot {E_A},{\rm{MLP}}\left( {\Tilde {{A_g}}} \right) \cdot {E_A}} \right]
$}
\vspace{-3pt}
\end{equation}
\noindent where $\Tilde {{A_a}} \in \mathbb{R}^{3D \times T}$, and $\Tilde {{A_g}} \in \mathbb{R}^{3D \times T}$ are fully integrated versions of acceleration and angular tokens, respectively. $\rm{MLP}(\cdot)$ refers to a multilayer perceptron network, and $[\cdot]$ denotes a concatenation operation. 
Subsequently, the attention module of $\rm{j}$-th transformer encoder layer is applied to capture multivariate correlations between temporal tokens, which is formulated as below:
\vspace{-3pt}
\begin{equation}
\scalebox{0.8}{
    $\Tilde {{A}}^j =\rm{SelfAttn}\left( {{\rm{query = }}\Tilde {{A}}^{j - 1}{\rm{ + }}\widetilde {E_A}^{j-1}{\rm{,  key = }}\Tilde {{A}}^{j - 1} + \widetilde {E_A}^{j-1},{\rm{ value = }}\Tilde {{A}}^{j - 1}} \right)$
}
\label{eq:selfattn}
\end{equation}

\noindent In this equation, $\Tilde {{A}}^j$ represents the updated tokens at layer $j$, while $\Tilde {{A}}^{j - 1}=[\Tilde {{A_a}}^{j-1},\Tilde {{A_g}}^{j-1}]$  and $\widetilde {E_A}^{j-1}$ are the tokens and positional embeddings from the previous layer, respectively. By applying adaptive scaling factors to the common base positional encodings $E_A$, iMoT becomes permutation-invariant to token orders, thus allowing it to tailor the positional encoding to specific characteristics of motion and rotation data. Staying in the flow of these benefits, the self-attention module, denoted as $\rm{SelfAttn}$ can better encode unique cross-modal interactions into contextual features.

\subsubsection{Adaptive Spatial Sync} 
\begin{figure}[t!]
\vspace{-4pt}
\centering
\includegraphics[width=0.34\textwidth] {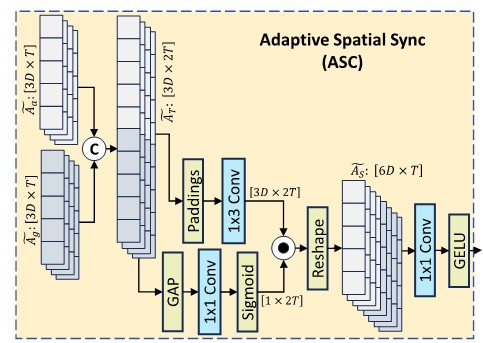} 
\vspace{-10pt}
\caption{Adaptive Spatial Sync.}
\label{fig:ASC}
\vspace{-15pt}
\end{figure}

Since contextual features are typically encoded in a temporal manner where cross-channel interactions at each time step are not considered, we propose an Adaptive Spatial Sync (ASC) module to compensate for the shortage of such spatial information. Placed at every residual connection, the ASC module can infuse fine-grained channel details directly along with two other information flows.
As shown in Fig.\ref{fig:ASC}, this infusion involves three stages: (1) an $1 \times 3$ convolution operated along 3D direction to extract cross-channel features from temporally concatenated tokens ${\widetilde {A}}_T \in \mathbb{R}^{3D \times 2T}$ for each time step; (2) a parallel branch with Global Average Pooling (GAP), $1 \times 1$ convolution, and sigmoid function to learn a channel weight vector from all sensor channels along the temporal axis $W = [w_1,\cdots, w_{2T}]$; and finally, (3) a Reshape operation followed by an $1 \times 1$ convolution with GELU to incorporate temporal interactions and then project the tokens back to their space.

\subsection{Decoder}
A given instantaneous velocity segment varies across individuals and is influenced by their unique gaits, giving rise to uncertainties in motion. To address this, we introduce $P$ motion pairs of query motion particles $\hat{v} \in \mathbb{R}^{P \times 2}$ and their query content features $C \in \mathbb{R}^{P \times T}$ into decoder layers, as illustrated in Fig.\ref{fig:imot}. Each pair represents a specific motion mode, collectively modeling motion uncertainties. By iteratively refining the query motion particles based on their updated content features for individual velocity segments, the model can quickly adapt to motion variability and effectively approximate the desired velocity segments.
\subsubsection{Query motion Particles}
The query motion particles, which characterize respective velocity, are represented as positional embeddings and dynamically updated according to the predicted velocity within each decoder layer as follows:
\vspace{-3pt}
\begin{equation}
    E^j_{\hat{v}} = \rm{MLP}(\rm{PE}(\widehat {{v}}^{j - 1}))
\end{equation}
\noindent where $\rm{PE}$ denotes the sinusoidal positional encoding operation, which is conditioned on velocity of the query particles ${\hat{v}}^{j - 1}$ in the previous layer $j-1$. The resulting learnable positional embeddings $E_{\hat{v}}^j$ are added to the content features $C_j$ (i.e., $C_0$ initialized with a zero matrix) and then passed into the $\rm{SelfAttn}$ module of the decoding layer $j$:
\vspace{-3pt}
\begin{equation}
\scalebox{0.8}{$
    C_{sa}^j = {\rm{SelfAttn}}\left( {{\rm{query}} = {C^{j - 1}} + E_{\widehat v}^j,{\rm{ key}} = {C^{j - 1}} + E_{\widehat v}^j,{\rm{ value}} = {C^{j - 1}}} \right)
$
}
\end{equation}
\noindent Here, $C_{sa}^j \in \mathbb{R}^{P \times T}$ represents the updated query content. To enable the pooling of cross-modal information from encoded context features, we concatenate positional embeddings with the updated content information as queries in the cross-attention module. This also allows us to decouple the contributions of content and position to the attention weights. To align with positional embeddings from the encoder, we learn an \rm{MLP} on the content information to generate a scaling vector for the positional embeddings in the decoder. Specifically, two cross-attention modules are employed singly for retrieving features regarding specific motion modes from both motion $\Tilde{A_a}$ and rotation tokens $\Tilde{A_g}$:
\vspace{-5pt}
\begin{equation}
    \scalebox{0.75}{
$
\begin{array}{l}
C_a^j = {\rm{SelfAttn}}\left( \begin{array}{l}
{\rm{query}} = \left[ {C_{sa}^j,{\rm{MLP}}\left( {{C^{j - 1}}} \right) \cdot E_{\widehat v}^j} \right],{\rm{ key}} = \left[ {{{\widetilde A}_a},{{\widetilde E}_a}} \right],\\
{\rm{ value}} = {\widetilde A_a}
\end{array} \right)\\
C_g^j = {\rm{SelfAttn}}\left( \begin{array}{l}
{\rm{query}} = \left[ {C_{sa}^j,{\rm{MLP}}\left( {{C^{j - 1}}} \right) \cdot E_{\widehat v}^j} \right],{\rm{ key}} = \left[ {{{\widetilde A}_g},{{\widetilde E}_g}} \right],{\rm{ }}\\
{\rm{value}} = {\widetilde A_g}
\end{array} \right)\\
{C^j} = {\rm{MLP}}\left( {\left[ {C_a^j,C_g^j} \right]} \right)
\end{array}
$
}
\end{equation}
\noindent where $C^j_a \in \mathbb{R}^{P \times T}$, $C^j_g \in \mathbb{R}^{P \times T}$, and $C^j \in \mathbb{R}^{P \times T}$  denote the acceleration-augmented content features, angular velocity-augmented content features, and the updated query content features for each query motion particle, respectively. Additionally, $\widetilde E_a \in \mathbb{R}^{3D \times T}$, $\widetilde E_g \in \mathbb{R}^{3D \times T}$ represent positional embeddings of motion and rotation tokens in $\widetilde E_A$.

\subsubsection{Particle Refinement}
In addition to external updates of positional embeddings, employing motion particles as learnable queries also enables internal, layer-by-layer updates, allowing the particle set to rapidly adapt to motion variability. For each decoding layer $j$, we utilize an $\rm{MLP}$ to predict relative velocity adjustments $\Delta \widehat v^j =[\Delta v^j_x, \Delta v^j_y]$ based on the updated content features. However, these $\rm{MLP}$s share the same parameters across layers to ensure consistent updates throughout the model.
\vspace{-5pt}
\begin{equation}
    \begin{array}{l}
\Delta \widehat v^j = {\rm{MLP}}\left( {\left[ {{C^j}} \right]} \right)\\
{\widehat v^{j + 1}} = {\widehat v^j} + \Delta \widehat v^j
\end{array}
\vspace{-3pt}
\end{equation}
\subsubsection{Dynamic Scoring Mechanism}
At the final layer, we start with the calculation of Euclidean distances between neighboring particles $\hat{v}$ and the ground-truth velocity particle $v_{GT} \in \mathbb{R}^{1 \times 2}$ to establish an inverse score list $S = [S_0,\cdots, S_{P-1}] \in \mathbb{R}^{1 \times P}$ that assigns higher scores to particles closer to $v_{GT}$. To optimize the particle set, we focus on making significant adjustments to distant particles while giving less attention to the closer ones when computing the mean particle $v_m \in \mathbb{R}^{1 \times 2}$. This strategy encourages a more compact distribution of particles around the desired velocity particle $v_{GT}$. The calculation is formalized  as follows:
\vspace{-3pt}
\begin{equation}
\begin{array}{l}
{S_p} = \frac{{{e^{ - d\left( {{{\widehat v}_p},{v_{GT}}} \right)}}}}{{\sum\limits_{i = 0}^{P - 1} {{e^{ - d\left( {{{\widehat v}_i},{v_{GT}}} \right)}}} }}\\
v_m = {\sum\limits_{i = 0}^{P - 1} {\left( {1 - {S_p}} \right)} ^\gamma }{\widehat v_p}\\
{J_{vel}} = \frac{1}{B}{\sum\limits_{b = 0}^{B - 1} {\left\| {v_{GT}^b - {v_m^b}} \right\|} ^2}
\end{array}
\vspace{-3pt}
\end{equation}
where $\gamma$ is the weighting factor largely controlling the influence of distant particles on the mean particle $v_m$, and $d(\cdot)$ represents the Euclidean distance operation. $J_{vel}$ is the mean square error loss between mean particles and ground-truth velocity particles over a batch of $B$ samples. To maintain stability during both training and testing, we further introduce an entropy loss function to maximize the entropy of the particle score list $S$:
\vspace{-5pt}
\begin{equation}
    {J_{ent}} = \frac{1}{{BP}}\sum\limits_{b = 0}^{B - 1} {\sum\limits_{p = 0}^{P - 1} { - S_p^b} } \log \left( {S_p^b + \varepsilon } \right)
    \vspace{-3pt}
\end{equation}
\noindent Here, $J_{ent}$ denotes the entropy loss function and $\epsilon$ is a small constant of $1e-10$ to stabilize the loss. This loss encourages a more uniform distribution of estimated particles around the ground-truth particles, allowing the use of average pooling of the estimated particles to approximate the desired velocities once training is complete. While the combined constraints of $J_{vel}$ and $J_{ent}$ could steer the proposed model to significantly minimize velocity discrepancies and adjust the particle distribution, we empirically find some instabilities during testing. 
Specifically, the absence of ground-truth particles for generating the score list necessitates the use of average pooling to approximate these particles, leading to insufficient fine-grained velocity segments and thereby degrading the quality of trajectory reconstruction.
To address this issue, we employ an $\rm{MLP}$ that attends to all aligned particles at the last layer, considering both $x$-, and $y$-directions, to directly learn a dynamic two-dimensional score list $S^d=[S^d_{1},\cdots,S^d_{P-1}] \in \mathbb{R}^{2 \times P}$:
\vspace{-4pt}
\begin{equation}
\begin{array}{l}
S^d = {\rm{MLP}}\left( {{{\widehat v}^{\rm T}}} \right)\\
v_m = {S^d} \cdot {\widehat v_p}
\end{array}
\vspace{-4pt}
\end{equation}
\noindent where ${{{\widehat v}^{\rm T}}}$ denotes transposed motion particles. This approach adaptively transforms the particle distribution and pools the mean particle accordingly during both training and testing, thereby eliminating the need for the entropy loss  $J_{ent}$. As a result, iMoT can be efficiently optimized using only the velocity loss $J_{vel}$.

\section{Experiment}
In this section, we empirically verify the effectiveness of our method in controllable and dynamic scenarios. 
\subsubsection{Dataset}
Four popular benchmark datasets are used for evaluation: RIDI \cite{yan2018ridi}, RoNIN \cite{herath2020ronin}, OxIOD \cite{chen2018oxiod}, and IDOL \cite{sun2021idol}. For controlled scenarios, RIDI and OxIOD are configured with predefined attachments, such as pocket, handheld, bag, body, and trolley, which are specified separately for each sequence. To provide end-users with greater freedom of movement in practice, IDOL and RoNIN are designed with dynamic motion contexts, where devices are naturally placed across all recorded sequences. Notably, RoNIN is the largest dataset, containing more than 40 hours of IMU data from 100 human subjects performing natural human motions.
\subsubsection{Evaluation Metric}
Four types of metrics are used for the quantitative trajectory evaluation:
\vspace{-1pt}
\begin{itemize}
     \setlength\itemsep{0.3pt}
    \item \textbf{Absolute Trajectory Error} (ATE) (m) is calculated as the average Root Mean Squared Error (RMSE) between the estimated and ground-truth trajectories as a whole.
    \item \textbf{Distance-Relative Trajectory Error} (D-RTE) (m), is calculated as the average RMSE between the estimated and the ground-truth over a fixed distance $d_r$ (\textit{i.e.}, 1 m).
    \item \textbf{Time-Relative Trajectory Error} (T-RTE) (m) is the average RMSE over a regular period $t_r$ (\textit{i.e.}, 1 minute).
    \item \textbf{Position Drift Error} (PDE) ($\%$) measures the drifting error at the final position relative to the traveled distance.
\end{itemize}
\vspace{-5pt}
\subsubsection{Implementation Details}
To initialize PSD module in the encoder, we implement $AvgPool_{k_2 \times k_1}$ with $k_1$ and $k_2$ set to $9$, $3$, respectively. For decoding, we empirically find that using a set of $P=128$ query motion particles representing $128$ motion modes is sufficient. Depending on the sampling rate of each dataset, the token dimension is set to $100$ for IMU sequences recorded at $100$ Hz and to $200$ for sequences recorded at $200$ Hz. The network, consisting of $N=2$ encoder layers and $M=2$ decoder layers, is trained with the learning rate of $1e-4$ and batch size of $B=128$ using Adam optimization. The training is performed with PyTorch version 2.4.0 on an H100 GPU with $80$ GB of memory.

\subsubsection{Ablation Study}

\begin{table*}[th!]
\centering
\caption{Ablation studies on the proposed module on RoNIN dataset.}
\vspace{-10pt}
\resizebox{1.\columnwidth}{!}{%
\begin{tabular}{ccccccccc}
\hline
\multirow{2}{*}{\textbf{Config}} &
  \multirow{2}{*}{\textbf{PSD}} &
  \multirow{2}{*}{\textbf{ASC}} &
  \multirow{2}{*}{\textbf{APE}} &
  \multirow{2}{*}{\textbf{Particles}} &
  \multirow{2}{*}{\textbf{DSM}} &
  \multicolumn{3}{c}{\textbf{Unseen}} \\ \cline{7-9} 
 &
   &
   &
   &
   &
   &
  \multicolumn{1}{l}{\textbf{ATE (m) $\downarrow$}} &
  \multicolumn{1}{l}{\textbf{T-RTE (m) $\downarrow$}} &
  \multicolumn{1}{l}{\textbf{D-TRE (m) $\downarrow$}} \\ \hline
\textit{(i)}    & -            & -            & -            & -            & -            & 6.48 & 5.53 & 0.41 \\
\textit{(ii)}   & -            & -            & -            & $\checkmark$ & -            & 7.10 & 4.61 & 0.42 \\
\textit{(iii)}  & -            & -            & $\checkmark$ & $\checkmark$ & -            & 6.33 & 4.65 & 0.41 \\
\textit{(iv)}   & -            & -            & -            & $\checkmark$ & $\checkmark$ & 6.13 & 4.61 & 0.38 \\
\textit{(v)}    & $\checkmark$ & -            & -            & $\checkmark$ & $\checkmark$ & 5.63 & 4.42 & 0.37 \\
\textit{(vi)}   & $\checkmark$ & -            & $\checkmark$ & $\checkmark$ & $\checkmark$ & 5.39 & 4.48 & 0.35 \\
\textit{(vii)}  & -            & -            & $\checkmark$ & $\checkmark$ & $\checkmark$ & 6.05 & 4.69 & 0.38 \\
\textit{(viii)} & $\checkmark$ & -            & -            & -            & -            & 5.58 & 4.41 & 0.37 \\
\textit{(ix)}   & -            & $\checkmark$ & -            & -            & -            & 6.00 & 4.70 & 0.39 \\
\textit{(x)}    & -            & -            & $\checkmark$ & -            & -            & 5.76 & 4.64 & 0.36 \\
\textit{(xi)}   & $\checkmark$ & $\checkmark$ & $\checkmark$ & -            & -            & 5.54 & 4.58 & 0.36 \\
\textit{(xii)}   & $\checkmark$ & -            & -            & $\checkmark$ & -            & 7.03 & 4.35 & 0.43 \\
\textit{(xiii)}  & $\checkmark$ & $\checkmark$ & -            & $\checkmark$ & $\checkmark$ & 5.41 & 4.53 & 0.37 \\
\textit{(xiv)} & -            & $\checkmark$ & $\checkmark$ & $\checkmark$ & $\checkmark$ & 5.79 & 4.52 & 0.37 \\
\textit{(xv)} &
  $\checkmark$ &
  $\checkmark$ &
  $\checkmark$ &
  $\checkmark$ &
  $\checkmark$ &
  \textbf{5.31} &
  \textbf{4.39} &
  \textbf{0.36} \\ \hline
\end{tabular}%
}
\label{tab:ablation}
\vspace{-5pt}
\end{table*}
As presented in Tab.\ref{tab:ablation}, we develop $14$ configurations on the largest RoNIN dataset, where all the proposed modules are progressively incorporated or removed, to meticulously examine both stand-alone and complementary contributions of the proposed modules.

\textbf{Progressive Series Decoupler.} To assess the isolated impact of PSD, we compare the performance between the baseline model $(i)$, where all proposed modules are excluded, and model $(viii)$ with only PSD enabled. The findings reveal considerable improvements across all error metrics when PSD is activated, especially a 13.89$\%$ reduction in ATE. Furthermore, the effectiveness of PSD is consistently observed even when jointly incorporated with other modules. For instance, there is an 8.16$\%$ ATE reduction between model $(iv)$ with PSD and model $(v)$ without PSD, and an 8.29$\%$ reduction between the full version $(xv)$ and model $(xiv)$, which is identical except for the absence of PSD. These consistent improvements across different configurations firmly verify the necessity of the additional information provided by PSD.

\textbf{Adaptive Positional Encoding.} To verify the stand-alone influence of APE, we collate the difference in performance between the baseline $(i)$ and model $(x)$ with only APE enabled. The improvements, particularly $\sim11.11\%$ in ATE reduction underscores APE's significant role in producing adaptive positional embeddings accounting for modal distinctions. Furthermore, consistent performance gains across various configurations pairs, \textit{e.g.}, $(iii)$ vs. $(ii)$, $(vi)$ vs. $(v)$, $(vii)$ vs. $(iv)$, and $(xiii)$ vs. $(xv)$ further validates the effectiveness of the proposed module with higher confidence.

\textbf{Adaptive Spatial Sync.} The ASC module is designed to compensate for the absence of cross-channel interactions. Although its contribution margin may appear smaller than others, particularly a 7.41$\%$ reduction in ATE when upgrading the baseline model $(i)$ to the configuration $(ix)$ with the inclusion of ASC, it has consistently delivered improvements both in standalone and complementary scenarios. Typically, the performance gains from the configuration $(vi)$ to the full version $(xv)$ further confirm the effectiveness of fusing spatial interactions, as evidenced by a reduction in ATE from 5.39 m to 5.31 m.

\textbf{Query Motion Particles \& Dynamic Scoring Mechanism.} We first examine the bare influences of manipulating query motion particles using the combined constraints of $J_{vel}$ and $J_{ent}$ to describe uncertainties in motion by comparing the baseline $(i)$ with model $(ii)$. Unlike other cases, this results in a deterioration of 0.62 m in ATE. Further investigation reveals that directly adding query particles to existing models, such as model $(viii)$ with PSD (which already outperformed the baseline $(i)$ by 13.89 $\%$), leads to worse performance. For instance, the upgraded model $(xii)$ performs 1.45 m worse than model $(viii)$ and 0.55 m worse than the baseline $(i)$. This can be attributed to the inefficacy of the above constraints in optimizing the particle set, which becomes severe as extra sources are introduced by PSD. A similar phenomenon is observed in configurations $(x)$ and $(iii)$, where introducing the particles exacerbates existing performance. 
However, these issues are significantly alleviated, and performance gains are achieved when the DSM module is introduced. Specifically, the configuration $(iv)$ with DSM surpasses all its previous versions, including $(ii)$ and $(i)$. Similar enhancements are witnessed in comparisons of $(iii)$ vs. $(vii)$ and $(xii)$ vs. $(v)$. 
 Besides, an intriguing observation is that $(viii)$ and $(x)$, which integrate only a single module (\textit{e.g.}, PSD or APE), outperform their counterparts $(v)$ and $(vii)$, which further incorporate both the query particles and DSM. This raises questions about the efficacy of these two components when combined. To investigate, an additional experiment was conducted, comparing the first three basic modules combined in $(xi)$ with the full version $(xv)$. Notably, the full version, which incorporates query particles and DSM,  achieves a further 3.99 $\%$ improvement in ATE over the combined model $(xi)$.
These findings underscore the critical importance of the query particle set in fully exploiting basic components, while also highlighting the necessity of a flexible optimization method like DSM to maximize their effectiveness.

\begin{figure}[t!]
\vspace{-4pt}
\centering
\includegraphics[width=0.35\textwidth] {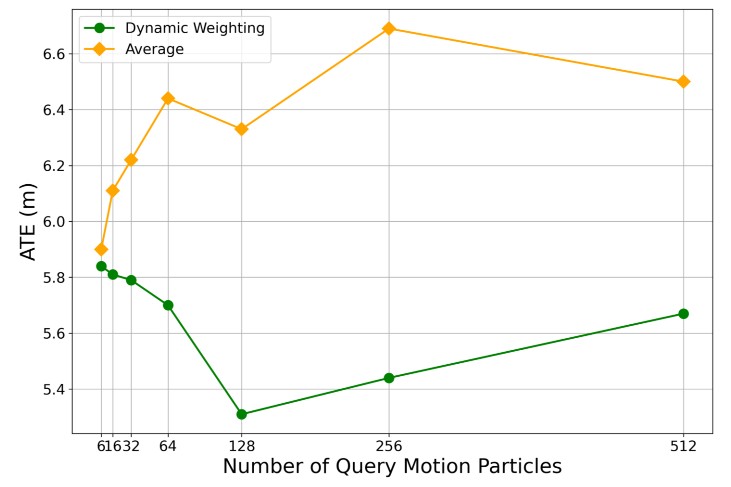} %
\vspace{-9pt}
\caption{Ablation Study on the number of query motion particles on RoNIN dataset.}
\label{fig:particle_num}
\vspace{-20pt}
\end{figure}

\begin{table*}[th!]
\centering
\caption{Overall Trajectory Prediction Evaluation. The best result is shown in bold font.}
\label{tab:sota}
\vspace{-10pt}
\resizebox{1.3\columnwidth}{!}{%
\begin{tabular}{cclccccccccc}
\hline
\multirow{2}{*}{Dataset} &
  \multirow{2}{*}{Test Subject} &
  \multicolumn{1}{c}{\multirow{2}{*}{Metric}} &
  \multicolumn{9}{c}{Method} \\ \cline{4-12} 
 &
   &
  \multicolumn{1}{c}{} &
  \multicolumn{1}{c}{SINS} &
  \multicolumn{1}{c}{PDR} &
  \multicolumn{1}{c}{RIDI} &
  \multicolumn{1}{c}{RoLSTM} &
  \multicolumn{1}{c}{RoTCN} &
  \multicolumn{1}{c}{RoResnet18} &
  \multicolumn{1}{c}{CTIN} &
  \multicolumn{1}{c}{TLIO} &
  \multicolumn{1}{c}{Ours} \\ \hline
\multirow{6}{*}{RIDI} &
  \multirow{3}{*}{Seen} &
  ATE (m) &
  \multicolumn{1}{c}{6.34} &
  \multicolumn{1}{c}{22.76} &
  \multicolumn{1}{c}{8.18} &
  \multicolumn{1}{c}{1.94} &
  2.56 &
  1.64 &
  1.69 &
  \textbf{1.67} &
  1.68 \\
 &
   &
  T-RTE (m) &
  \multicolumn{1}{c}{8.13} &
  \multicolumn{1}{c}{24.89} &
  \multicolumn{1}{c}{9.34} &
  \multicolumn{1}{c}{2.60} &
  2.81 &
  1.93 &
  2.03 &
  2.00 &
  \textbf{1.91} \\
 &
   &
  D-RTE (m) &
  \multicolumn{1}{c}{0.52} &
  \multicolumn{1}{c}{1.39} &
  \multicolumn{1}{c}{0.97} &
  \multicolumn{1}{c}{0.27} &
  0.27 &
  \textbf{0.21} &
  0.27 &
  \textbf{0.21} &
  \textbf{0.21} \\ \cline{2-12} 
 &
  \multirow{3}{*}{Unseen} &
  ATE (m) &
  \multicolumn{1}{c}{4.62} &
  \multicolumn{1}{c}{20.56} &
  \multicolumn{1}{c}{8.18} &
  \multicolumn{1}{c}{2.24} &
  2.15 &
  1.76 &
  2.15 &
  1.85 &
  \textbf{1.49} \\
 &
   &
  T-RTE (m) &
  \multicolumn{1}{c}{4.58} &
  \multicolumn{1}{c}{31.17} &
  \multicolumn{1}{c}{10.51} &
  \multicolumn{1}{c}{2.70} &
  1.95 &
  1.71 &
  1.84 &
  1.82 &
  \textbf{1.33} \\
 &
   &
  D-RTE (m) &
  \multicolumn{1}{c}{0.36} &
  \multicolumn{1}{c}{1.19} &
  \multicolumn{1}{c}{1.09} &
  \multicolumn{1}{c}{0.31} &
  0.23 &
  0.21 &
  0.27 &
  0.23 &
  \textbf{0.20} \\ \hline
\multirow{6}{*}{RoNIN} &
  \multirow{3}{*}{Seen} &
  ATE (m) &
  7.89 &
  26.64 &
  16.82 &
  4.14 &
  5.81 &
  4.13 &
  5.54 &
  4.41 &
  \textbf{3.78} \\
 &
   &
  T-RTE (m) &
  5.30 &
  23.82 &
  19.50 &
  2.83 &
  3.56 &
  2.81 &
  3.26 &
  2.82 &
  \textbf{2.68} \\
 &
   &
  D-RTE (m) &
  0.42 &
  0.98 &
  4.99 &
  0.29 &
  0.37 &
  \textbf{0.26} &
  0.34 &
  \textbf{0.26} &
  \textbf{0.26} \\ \cline{2-12} 
 &
  \multirow{3}{*}{Unseen} &
  ATE (m) &
  7.62 &
  23.49 &
  15.75 &
  6.90 &
  7.19 &
  5.95 &
  6.89 &
  6.77 &
  \textbf{5.31} \\
 &
   &
  T-RTE (m) &
  5.12 &
  23.07 &
  19.13 &
  4.46 &
  5.15 &
  4.53 &
  4.95 &
  4.69 &
  \textbf{4.39} \\
 &
   &
  D-RTE (m) &
  0.43 &
  1.00 &
  5.37 &
  0.42 &
  0.47 &
  \textbf{0.36} &
  0.43 &
  0.39 &
  \textbf{0.36} \\ \hline
\multirow{6}{*}{OxIOD} &
  \multirow{3}{*}{Seen} &
  ATE (m) &
  15.36 &
  9.78 &
  3.78 &
  2.00 &
  2.12 &
  2.45 &
  6.71 &
  2.51 &
  \textbf{1.86} \\
 &
   &
  T-RTE (m) &
  11.02 &
  8.51 &
  3.99 &
  1.93 &
  1.92 &
  1.02 &
  2.31 &
  1.16 &
  \textbf{0.94} \\
 &
   &
  D-RTE (m) &
  0.96 &
  1.16 &
  2.30 &
  0.62 &
  0.62 &
  0.22 &
  0.33 &
  0.22 &
  \textbf{0.21} \\ \cline{2-12} 
 &
  \multirow{3}{*}{Unseen} &
  ATE (m) &
  13.90 &
  17.72 &
  7.16 &
  2.03 &
  2.00 &
  1.06 &
  2.34 &
  1.03 &
  \textbf{0.90} \\
 &
   &
  T-RTE (m) &
  10.51 &
  17.21 &
  7.65 &
  1.40 &
  1.35 &
  \textbf{1.15} &
  1.53 &
  1.26 &
  1.32 \\
 &
   &
  D-RTE (m) &
  0.89 &
  1.10 &
  2.62 &
  0.62 &
  0.61 &
  \textbf{0.21} &
  0.28 &
  0.22 &
  0.22 \\ \hline
\multirow{6}{*}{IDOL} &
  \multirow{3}{*}{Seen} &
  ATE (m) &
  21.54 &
  18.44 &
  9.79 &
  3.68 &
  4.66 &
  2.70 &
  3.15 &
  2.90 &
  \textbf{2.22} \\
 &
   &
  T-RTE (m) &
  14.93 &
  14.53 &
  7.97 &
  3.78 &
  5.58 &
  2.45 &
  3.05 &
  2.63 &
  \textbf{1.86} \\
 &
   &
  D-RTE (m) &
  1.07 &
  1.14 &
  0.97 &
  0.43 &
  0.53 &
  0.27 &
  0.34 &
  0.30 &
  \textbf{0.24} \\ \cline{2-12} 
 &
  \multirow{3}{*}{Unseen} &
  ATE (m) &
  20.34 &
  16.83 &
  9.54 &
  4.34 &
  5.03 &
  3.32 &
  3.70 &
  3.36 &
  \textbf{3.00} \\
 &
   &
  T-RTE (m) &
  18.48 &
  15.67 &
  9.07 &
  5.15 &
  6.15 &
  3.37 &
  4.12 &
  3.50 &
  \textbf{2.85} \\
 &
   &
  D-RTE (m) &
  1.36 &
  1.31 &
  1.04 &
  0.47 &
  0.57 &
  0.32 &
  0.38 &
  0.33 &
  \textbf{0.28} \\ \hline
\end{tabular}%
}
\vspace{-5pt}
\end{table*}

\begin{figure*}[th!]
\vspace{-4pt}
\includegraphics[width=0.73\textwidth] {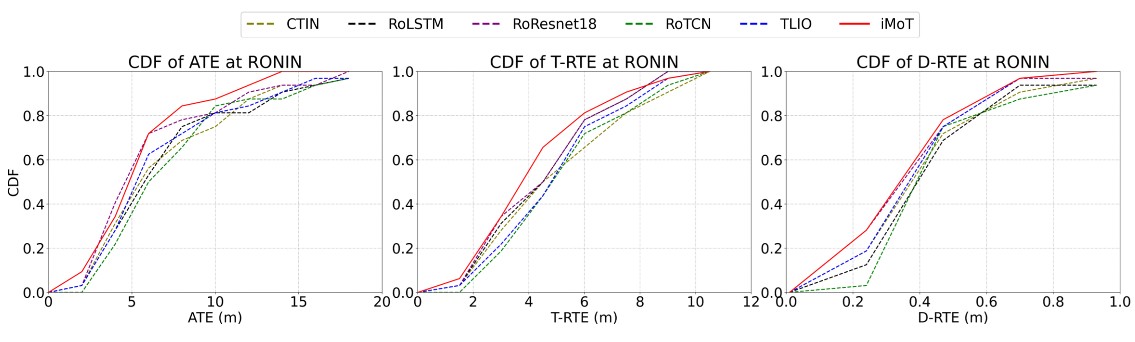} %
\includegraphics[width=.28\textwidth]{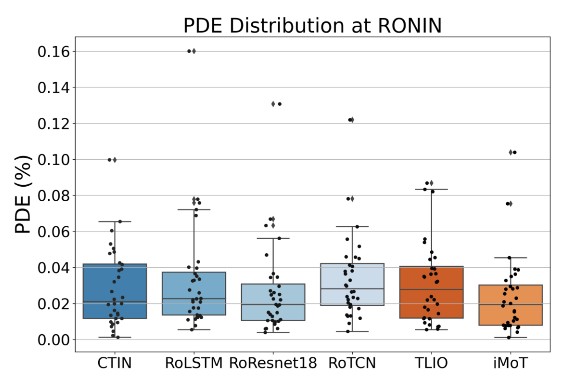}
\vspace{-20pt}
\caption{Cumulative Error Distributions (CDF) with three types of metric types, and boxplot of PDE on RoNIN dataset.}
\label{fig:cdf}
\vspace{-12pt}
\end{figure*}
\textbf{Particle Number.} The number of query motion particles plays a crucial role in extracting cross-modal interactions from context features and describing motion uncertainties thereafter. As shown in Fig.\ref{fig:particle_num}, using $J_{vel}$ and $J_{ent}$ with average pooling after training cannot tolerate a higher number of particles, limiting the model's capacity. However, the introduction of DSM allows the model to effectively accommodate more particles, enhancing its ability to represent motion uncertainties. Our experiments reveal that 128 particles are optimal for modeling uncertainties across different velocity segments in RoNIN. Fewer than 128 particles may fail to cover enough motion modes for accurate velocity synthesis, while more than 128 particles could introduce redundancy, potentially diluting cross-modal information with unnecessary motion modes through their positional embeddings.
\vspace{-4pt}
\subsubsection{State-of-The-Art Performance}
The results presented in Tab. \ref{tab:sota} exhibit a detailed evaluation of trajectory errors for various odometry methods across four benchmark datasets: RIDI, RoNIN, OxIOD, and IDOL, using three standard metrics: ATE, T-RTE, and D-RTE. Throughout the experiment, our proposed method consistently outperforms other SoTA approaches, particularly in its ability to generalize to unseen subjects. For example, in dynamic scenarios from the RoNIN dataset, where recording devices are freely held during movement, our method achieves an ATE of 5.31 m for unseen subjects, significantly outperforming RIDI (15.75 m) by 66.29$\%$, and other robust methods that account for motion uncertainties, such as CTIN (6.89 m) and TLIO (6.77 m), by 22.93$\%$ and 21.57$\%$, respectively. This trend of superior performance is consistent across all evaluated datasets in terms of D-RTE and T-RTE as well. For instance, on the IDOL dataset, our model demonstrated a 15.43$\%$ improvement in T-RTE and a 12.50$\%$ improvement in D-RTE compared to the second-best model, RoResnet18. These results can be attributed to our method's distinct advantage in effectively modeling motion uncertainties with learnable query motion particles. While other methods show a significant performance decline in unseen dynamic scenarios, the minimal error increase between seen and unseen subjects in our method highlights its generalization capability.

For stability verification, we further visualize the detailed metrics over the entire RoNIN dataset. 
As illustrated in Fig. \ref{fig:cdf}, the cumulative distribution function (CDF) of the trajectory errors clearly shows that the proposed method consistently outperforms the others, as indicated by the uppermost positions of the red curve. Specifically, in 80$\%$ of the cases predicted by our method, the ATE remains below approximately 6.5 m, denoted as $P(X<6.5)=0.8$. In contrast, other methods only achieve an ATE upper bound of around 10 m for 80$\%$ of the examples. Furthermore, the network also exhibits the lowest position drift error (PDE $\%$) with the highest confidence and fewest outliers, further highlighting its robustness and precision in trajectory prediction.
\vspace{-8pt}
\section{Conclusion}
\vspace{-0.5pt}
This paper introduces iMoT, an innovative transformer architecture designed to capture and model motion uncertainties within instantaneous velocity segments. Extensive experimental results demonstrated that PSD greatly enhances the encoding of complex temporal patterns, while the manipulation of the query particle set effectively learns and represents various motion modes, accounting for motion variability. Our approach not only improves trajectory reconstruction quality but also exhibits robust generalization across a wide range of dynamic scenarios, setting a new SoTA standard in handling motion uncertainty for odometry tasks.
\section*{Acknowledgement}
This publication is part of the project MOSAIC: enhancement of MicrOfluidic Sensing with deep symbolic Artificial IntelligenCe with file number 19985 of the research programme Open Technology Programme which is (partly) financed by the Dutch Research Council (NWO).

This work made use of the Dutch national e-infrastructure with the support of the SURF Cooperative using grant no. EINF-6216.

\bibliography{aaai25}

\newpage


\appendix
\appendixpage 
\section{Additional Details on Baseline Models and Metrics}

 \textit{Trajectory reconstruction} is performed by doing integration of predicted velocity segments. The major metric used to evaluate the accuracy of positioning is a Root Mean Squared Error (RMSE) with various definitions of the inside estimation error: ${\rm{RMSE}} = \sqrt {\frac{1}{T}\sum\limits_{t = 0}^{T - 1} {\left\| {{\rm{E}_t}\left( {{x_t},{{\widetilde x}_t}} \right)} \right\|} }$, where ${{\rm{E}_t}\left( {{x_t},{{\widetilde x}_t}} \right)}$ represents an estimation error at timestamp $t$ between a position $x_t$ in the
ground truth trajectory and its corresponding one ${\widetilde x}_t$ in the predicted path. Although standard metrics are described in the paper, all of them are additionally formulated as below:

\begin{itemize}
    \item \textbf{Absolute Trajectory Error (ATE)} is RMSE of squared Euclidian error distances ${\rm{E}_t}(x_t, {\widetilde x}_t) = {d(x_t - {\widetilde x}_t)}^2$
    \item \textbf{Time-Relative Trajectory Error (T-RTE)} is the time-normalized RMSE of average errors ${{\rm{E}}_{\rm{t}}}\left( {{x_t},{{\widetilde x}_t}} \right) = {d\left(\left( {{x_{t + {t_r}}} - {x_t}} \right) - \left( {{{\widetilde x}_{t + {t_r}}} - {{\widetilde x}_t}} \right)\right)}^2$.
    \item \textbf{Distance-Relative Trajectory Error (D-RTE)} is the distance-normalized RMSE of average errors ${{\rm{E}}_{\rm{t}}}\left( {{x_t},{{\widetilde x}_t}} \right) = {d\left(\left( {{x_{t + {t_d}}} - {x_t}} \right) - \left( {{{\widetilde x}_{t + {t_d}}} - {{\widetilde x}_t}} \right)\right)}^2$ where $t_d$  is the time required to travel a standard distance $d_r$ of 1 m.
    \item \textbf{Position Drift Error (PDE)} measures the drift in the final position relative to the total distance traveled. ${\rm{PDE}} = \frac{{d\left( {{x_{_{T - 1}}} - {{\widetilde x}_{T - 1}}} \right)}}{{\sum\limits_{t = 0}^{T - 1} {d\left( {{x_t} - {x_t}} \right)} }}$
\end{itemize}

\textit{Baseline Methods}:
\begin{itemize}
    \item \textbf{Strap-down Inertial Navigation System (SINS)}: The subject's position is derived from double integration of linear accelerations, after subtracting Earth's gravity. This process involves rotating the accelerations from the body frame to the navigation frame using device orientation data, followed by performing a double integral on the rotated accelerations to obtain position estimates \cite{savage1998strapdown}.

    \item \textbf{Pedestrian Dead Reckoning (PDR)}: We employ an open-source step-counting algorithm \cite{tian2015enhanced} to detect foot-steps and update the position with each step, following the device's heading direction. A fixed stride length of 0.67 m/step is assumed.

    \item \textbf{Robust IMU Double Integration (RIDI)}: Using the original implementation \cite{yan2018ridi}, we train separate models for each device attachment in the RIDI and OxIOD datasets. For the other datasets, where data acquisition involved mixed attachments, we train a unified model for each dataset.

    \item \textbf{Robust Neural Inertial Navigation (RoNIN)}: We evaluate the performance using the original implementation of three types of neural networks (i.e., RoLSTM, RoResNet18, and RoTCN) as outlined by  RoNIN work\cite{herath2020ronin}.

    \item \textbf{Tight Learned Inertial Odometry (TLIO)}: We utilize the implementation of RoResNet18 and employ a joint training scheme that includes two types of losses: velocity loss and covariance loss \cite{liu2020tlio}.

    \item \textbf{Robust Contextual Transformer Network for Inertial Navigation (CTIN)}: Similar to TLIO, the transformer model \cite{rao2022ctin} is first trained with a velocity loss to warm up the network, followed by combined training with covariance loss to enhance performance.
    
\end{itemize}

\section {Explanation for the inferiority in State-of-The-Art Performance Comparison}

As shown in Tab.\ref{tab:sota}, RoResNet18 demonstrates slightly better T-RTE and comparable D-RTE results relative to our method. This can be further understood by examining the T-RTE plot, particularly the PDE plot in Fig.\ref{fig:cdf_OxIOD}.

As shown in the T-RTE plot, our proposed method achieves the best cumulative distribution function (CDF) of approximately 78$\%$ of testing samples with T-RTE errors less than 1 m ($P(X < 1) = \sim 78 \%$). However, we have a small number of samples with errors close to 3 m in the remaining 22$\%$, compared to RoResNet18’s maximum errors of around 2 m, which deviates our average T-RTE farther and thus results in a higher average error in T-RTE for our method. The PDE plot further illustrates this, which shows our method’s lower mean error but a higher percentile of outliers compared to RoResNet18.

Since OxIOD is specified with attachment settings individually predefined for each sequence, such as pocket, handheld, bag, body, and trolley, thus exhibiting reduced motion variability, the query motion particles in our method may not be fully utilized. In contrast, for dynamic attachments with significantly higher motion uncertainty, where devices are naturally placed across all recorded sequences to simulate real-world scenarios in RoNIN and IDOL, our method, which maximizes the use of query motion particles to represent motion uncertainties, consistently outperforms competing methods by a substantial margin.

\section{Qualitative Assessment}

In this section, we analyze two scenarios where the regression quality of velocity segments reveals the superior performance of our proposed model over the second-best method, RoResnet18.

\subsection{Lower velocity Loss, Better Trajectory}
\begin{figure*}[th!]
\vspace{-4pt}
\includegraphics[width=\textwidth]{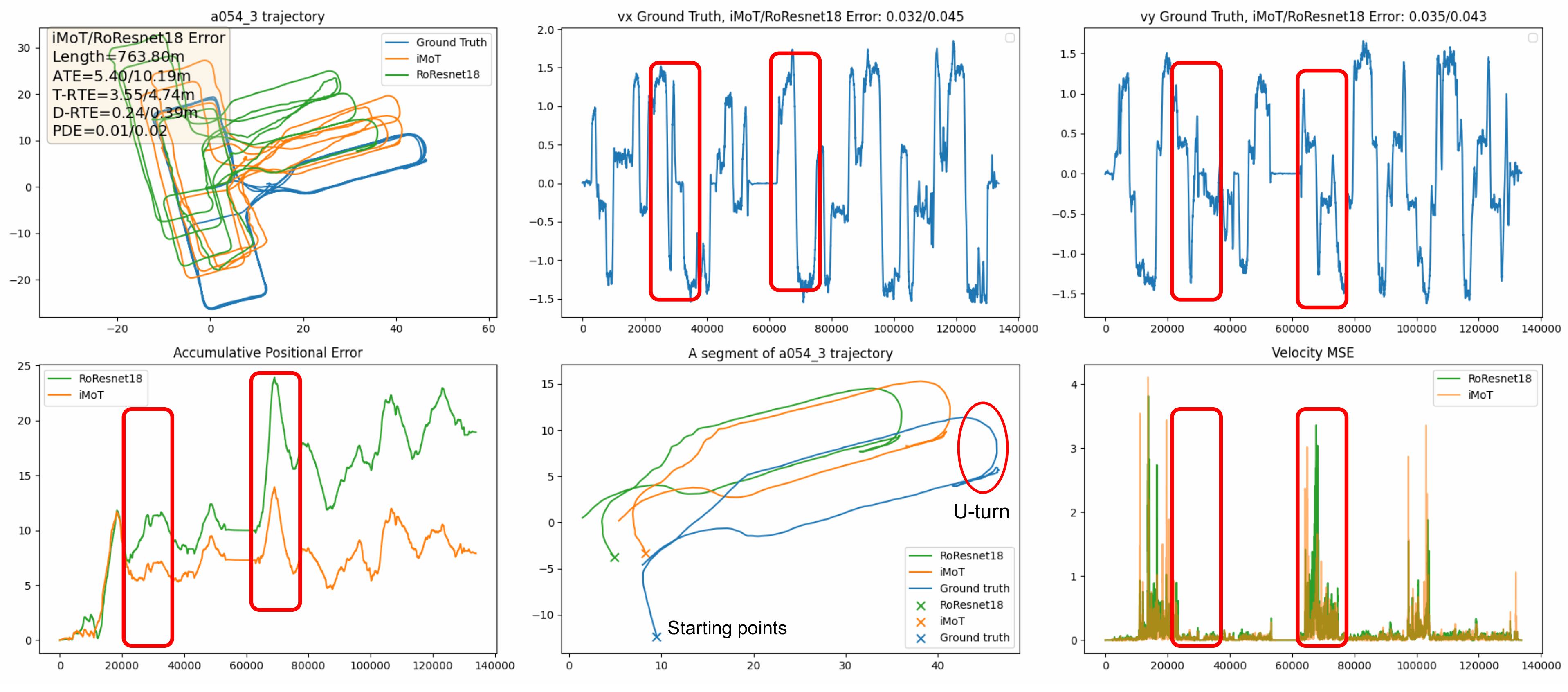}
\caption{
Quantitative Assessment between iMoT and RoResnet18 for Subject 54 on Trajectory 3 in the RoNIN Dataset. Best viewed in color and zoomed in for details.}
\label{fig:traj_analysis_54}
\end{figure*}

As illustrated in Fig.\ref{fig:traj_analysis_54}, our model achieves noticeably lower velocity errors in both the $x$ and $y$ directions, straightforwardly leading to more accurate trajectory reconstruction. The top-left plot highlights a significant improvement, with our method achieving a 47.01$\%$ reduction in ATE compared to RoResnet18. Further analysis of the bottom-left plot, which shows cumulative positional errors over time, indicates that both methods initially perform similarly from the time step 0 to 20k. However, from 20k to approximately 30k (around 50 seconds with a sampling rate of 200 Hz), iMoT begins to show a slightly different trend with declining errors. This leads to a much lower error margin for subsequent segments later on, particularly a 9 m reduction in error between the plateau and peak during the period from the time step 60k to 80k.

To gain more insights, we examine such periods in the top plots, which depict velocity progression along the $x$- and $y$-axes. In the first red window, we observe two primary patterns: an $x$-axis uptrend and a $y$-axis downtrend, indicating a downturn curve, followed by $x$- and $y$-axis downtrends that represent a linear backward movement, collectively illustrating a U-turn. This maneuver is a critical part of the trajectory, as it introduces significant confusing fluctuations for accurate reconstruction compared to linear segments. Errors in velocity during linear segments may just locally distort the trajectory, but errors during rotational segments, such as a 5-degree deviation, can cause the reconstructed trajectory to diverge significantly from the ground truth, compounding positional errors in subsequent segments.

To accurately execute a U-turn, precise rotational information is crucial. Our model shows smaller directional velocity errors compared to RoResnet18 during these phases, as depicted in the bottom-right plot. This likely contributes to the substantial reduction in errors for subsequent segments. The improved accuracy in U-turn execution is further illustrated in the bottom-middle plot, where iMoT performs a more accurate U-turn compared to RoResnet18, which performs an early U-turn and consequently deviates significantly from the ground truth.

\subsection{Higher velocity Loss, Still Better Trajectory}

\begin{figure*}[th!]
\vspace{-4pt}
\includegraphics[width=\textwidth]{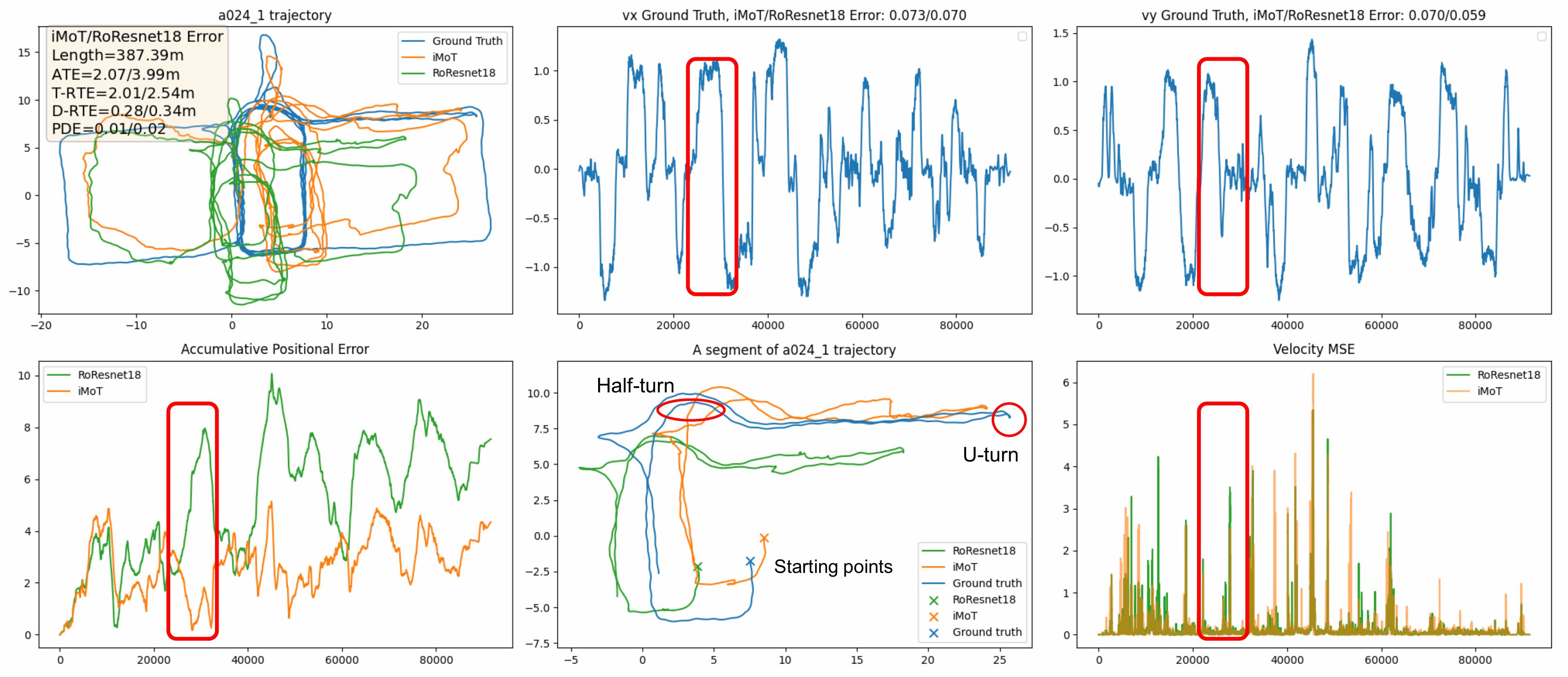}
\caption{
Quantitative Assessment between iMoT and RoResnet18 for Subject 24 on Trajectory 1 in the RoNIN Dataset. Best viewed in color and zoomed in for details.}
\label{fig:traj_analysis_24}
\end{figure*}

In Fig.\ref{fig:traj_analysis_24}, despite our method initially producing higher velocity errors compared to RoResnet18—specifically, 0.073 m vs. 0.070 m for the $x$-axis, and 0.070 m vs. 0.059 m for the $y$-axis—it achieves a significantly better trajectory reconstruction, with a 48.12$\%$ improvement in ATE (2.07 m vs. 3.99 m). As illustrated in the bottom-left plot, iMoT initially incurs higher positional errors from the start up to the time step 16k. However, it then demonstrates a substantial reduction in errors, particularly between the 16k and 39k time steps, where the trend reverses in our favor.

To better understand this, we focus on the period between the time step 20k and 39k, marked by a notable downtrend in iMoT's errors. Within the red-highlighted window of the velocity plots on the top, two key turning events are apparent:

\begin{itemize}
    \item The $x$-axis shows a transition from a plateau beginning at 20k, accompanied by the $y$-axis rising to a peak around 23k, followed by a rapid descent and a period of stabilization. This pattern suggests the subject was initially moving straight, made a right uphill turn, and then briefly went downhill, indicating a rightward straight movement.
    \item The subsequent decline in the $x$-axis, alongside fluctuations near zero on the $y$-axis, indicates that the subject turned back to the left.
\end{itemize}

During these events, RoResnet18 produced higher errors, as shown in the bottom-right plot, which tracks the mean square error of velocity over time. The bottom-middle plots reveal that RoResnet18 executed a premature right turn, deviating from the ground truth, followed by an incomplete U-turn that shortened its trajectory, misaligning it further from the actual path. 

In contrast, our method accurately captured these turning events, even though it deviated more from the ground truth in the beginning. Additionally, the bottom-middle plot shows that our method, while producing higher velocity errors in linear segments (reflected in some fluctuations during the rightward movement), ultimately provided robust trajectory reconstruction. This robustness is due to our model's superior handling of rotational information during turning events, which outweighs the initial higher errors in linear segments.\\[5pt]
\noindent \textit{Overall}, we attribute these advantages to the synergistic effect of our proposed modules. In particular, the \textit{Progressive Series Decoupler} (PSD) plays a crucial role by emphasizing critical motion events, enabling iMoT to better manage these events through the progressive decomposition of seasonal and trend-cycle signals. Furthermore, the enhanced flexibility offered by the \textit{learnable query particle set}, coupled with its \textit{dual updating} approach, both external and internal, and the \textit{Dynamic Scoring Mechanism} (DSM) for velocity synthesis, further strengthens our method. Together, these components empower our approach to rapidly capture motion events within IMU sequences and effectively adapt to them by refining potential motion modes through continuous updates (query particle set + dual updating scheme + DSM).

\section{Long-term Stability Verification}

\begin{figure*}[t!]
\vspace{-4pt}
\includegraphics[width=0.73\textwidth] {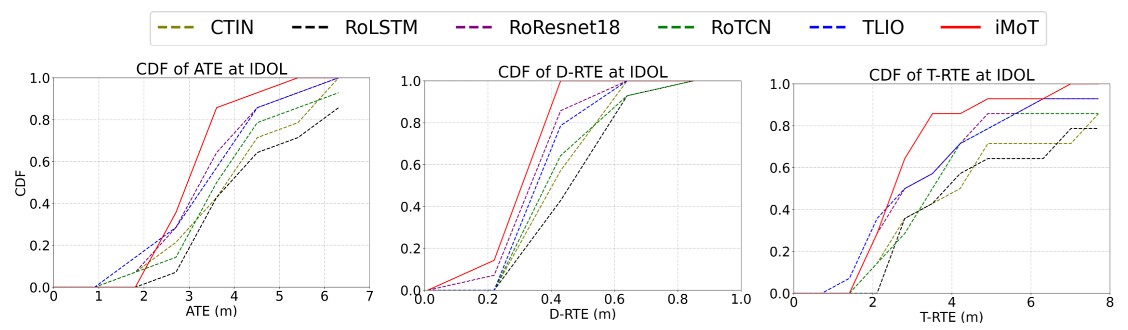} %
\includegraphics[width=.28\textwidth]{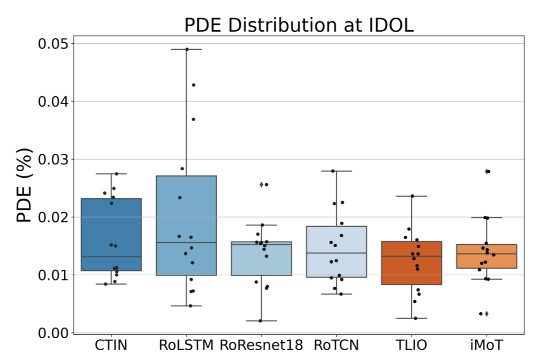}

\caption{Cumulative Error Distributions (CDF) with three types of metric types, and boxplot of PDE on IDOL dataset.}
\label{fig:cdf_IDOL}

\end{figure*}

\begin{figure*}[t!]
\vspace{-4pt}
\includegraphics[width=0.73\textwidth] {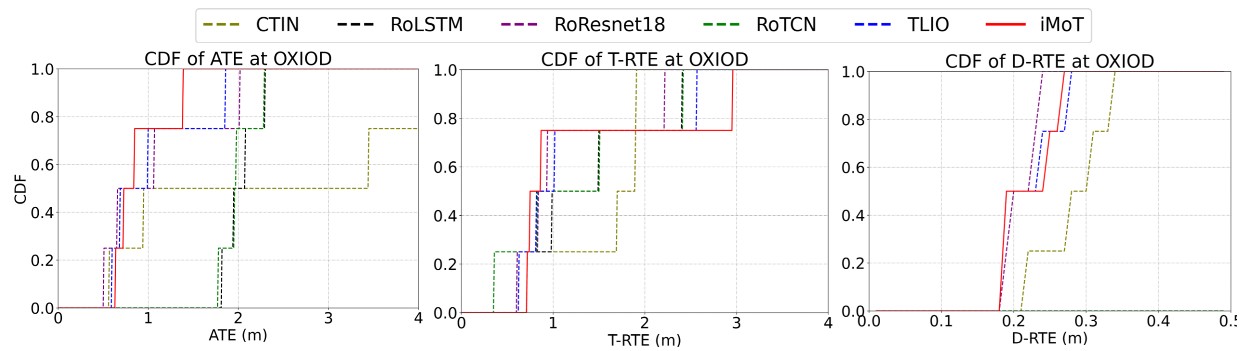} %
\includegraphics[width=.28\textwidth]{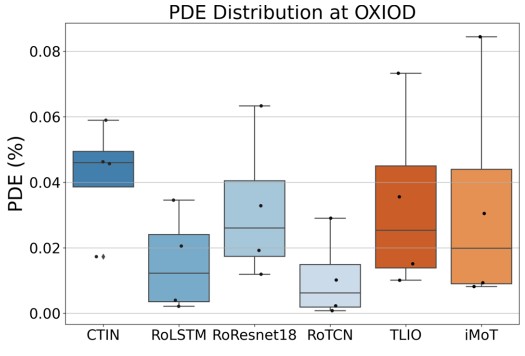}

\caption{Cumulative Error Distributions (CDF) with three types of metric types, and boxplot of PDE on OxIOD dataset.}
\label{fig:cdf_OxIOD}

\end{figure*}

\begin{figure*}[t!]
\vspace{-4pt}
\includegraphics[width=0.73\textwidth] {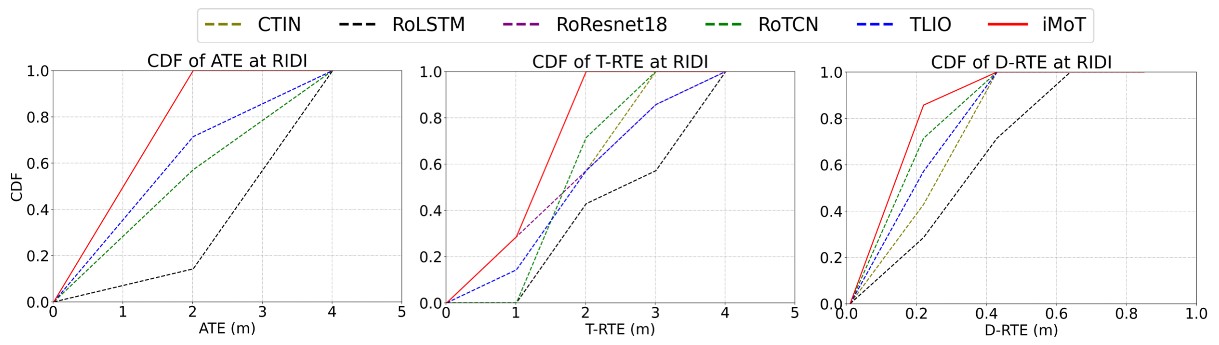} %
\includegraphics[width=.28\textwidth]{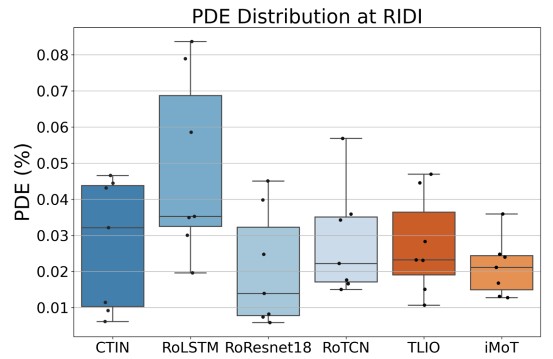}

\caption{Cumulative Error Distributions (CDF) with three types of metric types, and boxplot of PDE on RIDI dataset.}
\label{fig:cdf_RIDI}

\end{figure*}
In this section, we further evaluate the proposed model using cumulative distribution function (CDF) plots for all standard metrics across three additional datasets: IDOL, OxIOD, and RIDI. These plots, presented in Fig.\ref{fig:cdf_IDOL} to Fig.\ref{fig:cdf_RIDI}, provide a comprehensive overview of our model's effectiveness in comparison to other methods.

\section{Trajectory Visualization}
\begin{figure*}[t!]
\vspace{-4pt}
\includegraphics[width=\textwidth] {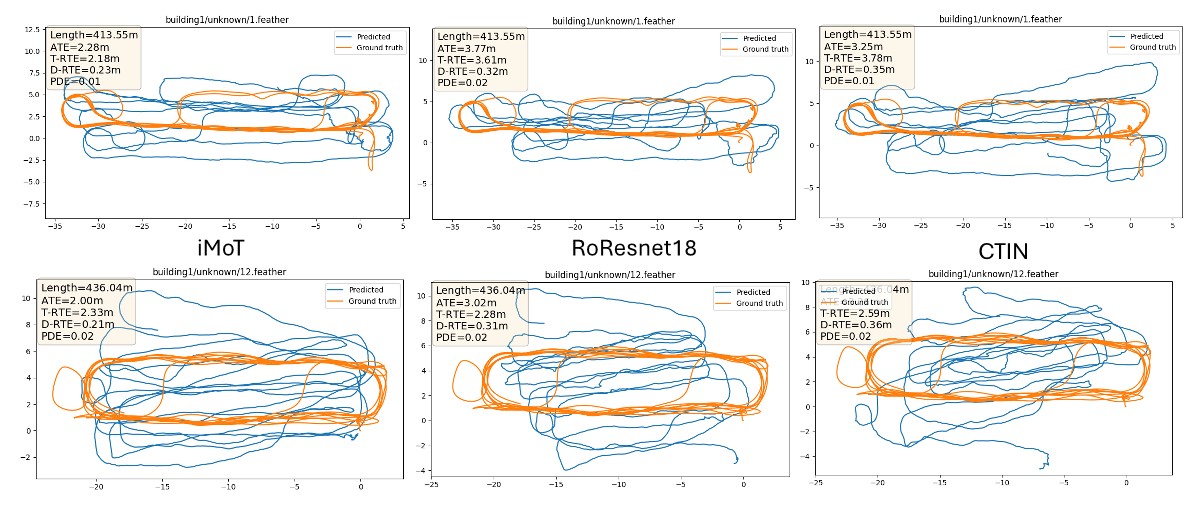}\\
\caption{Selected visualizations of trajectories for iMoT, CTIN, and RoResNet18 on the IDOL. }
\label{fig:traj_analysis_IDOL}
\end{figure*}

\begin{figure*}[t!]
\vspace{-4pt}
\includegraphics[width=\textwidth] {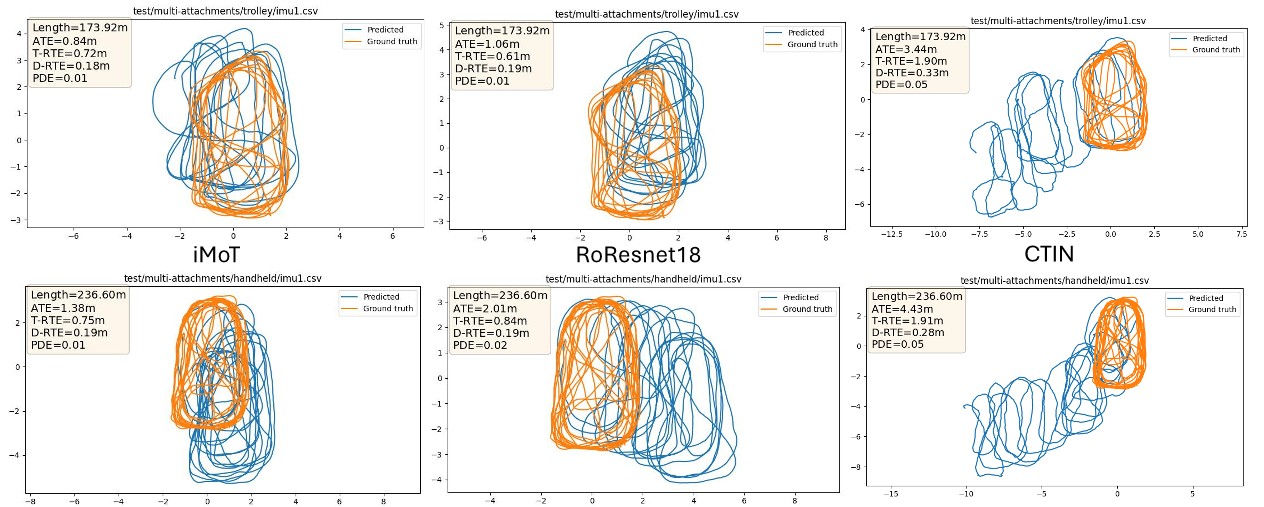}\\
\caption{Selected visualizations of trajectories for iMoT, CTIN, and RoResNet18 on the OxIOD.}
\label{fig:traj_analysis_OXIOD}
\end{figure*}

\begin{figure*}[t!]
\vspace{-4pt}
\includegraphics[width=\textwidth] {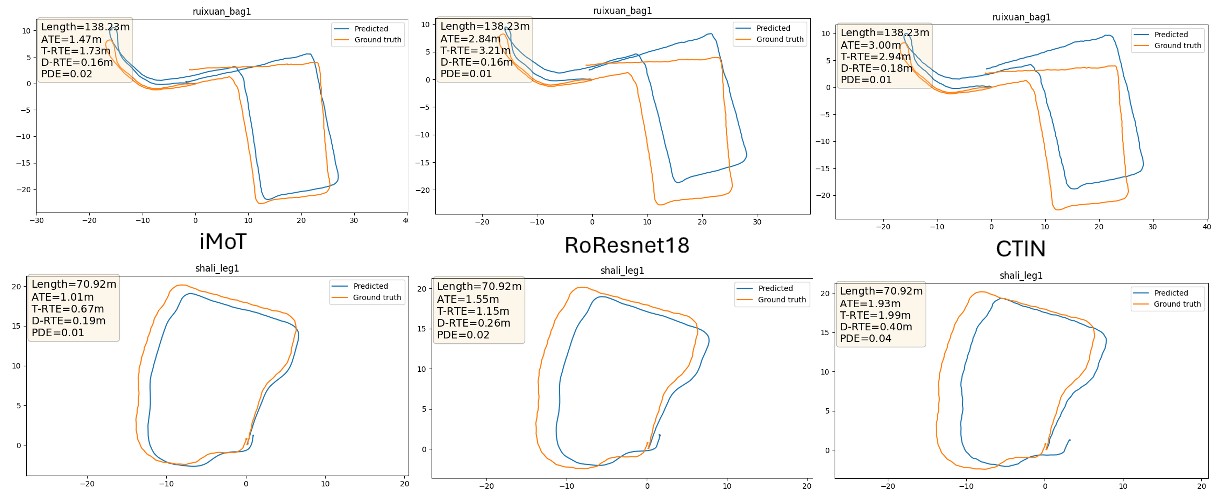}\\
\caption{Selected visualizations of trajectories for iMoT, CTIN, and RoResNet18 on the RIDI. }
\label{fig:traj_analysis_RIDI}
\end{figure*}

\begin{figure*}[t!]
\vspace{-4pt}
\includegraphics[width=\textwidth] {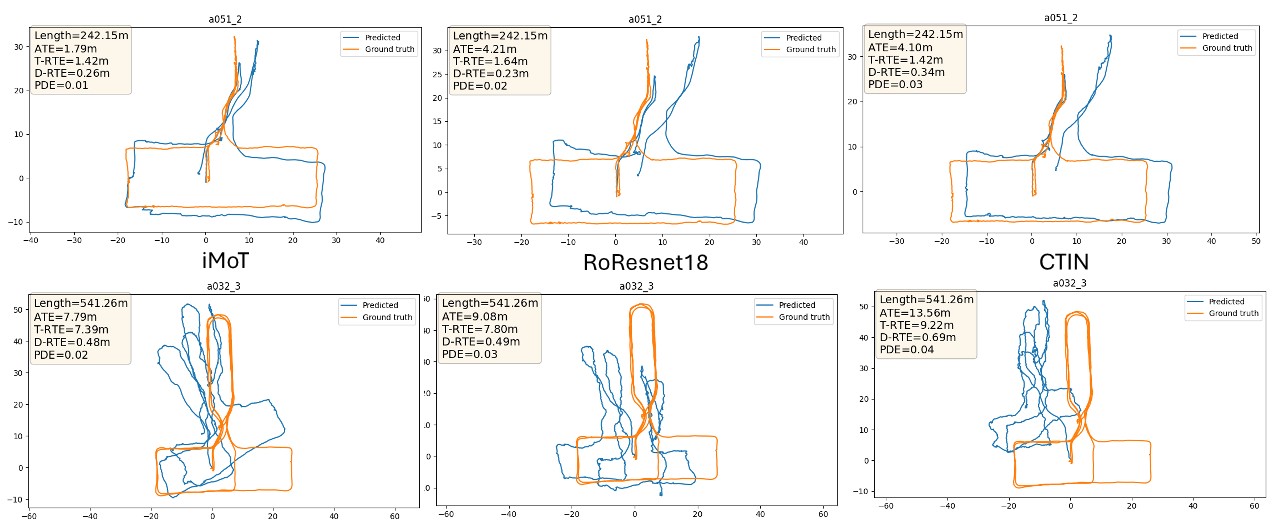}\\
\caption{Selected visualizations of trajectories for iMoT, CTIN, and RoResNet18 on the RoNIN.}
\label{fig:traj_analysis_RoNIN}
\end{figure*}

In this section, we present additional selected trajectories, depicted from Figs. \ref{fig:traj_analysis_IDOL} to Fig.\ref{fig:traj_analysis_RoNIN}, which are generated by unseen subjects across four datasets: IDOL, OxIOD, RIDI, and RoNIN. Each plot includes all standard metrics, highlighted in yellow boxes positioned in the top-left corner.

\section{Computation Cost}
For IMU measurements sampled at 200 Hz, our proposed model contains approximately 14.49M parameters, around three times and seven times more than RoResnet (4.63M parameters), and RoTCN (2.03M parameters). However, our model operates at only 7.79 GFLOPS per second, which is notably 1.37, and 25.38 GFLOPS far lower than that of RoResnet and RoTCN, respectively.

Importantly, all architectural configurations of the comparison models were selectively optimized in their respective papers for the desired accuracy, and we have preserved their original configurations and functionality to ensure a fair comparison. Although our model has a larger footprint, it stands fully suitable for integration into embedded devices, especially given its computational efficiency.

\begin{table}[]
\caption{Computation Cost and Footprint for IMU measurements sampled at 200 Hz}
\centering
\resizebox{\columnwidth}{!}{%
\begin{tabular}{lcc}
\hline
\multicolumn{1}{l}{\textbf{Model}} &
  \textbf{\begin{tabular}[c]{@{}c@{}}No. of Parameters\\ $1 \times 10^6$\end{tabular}} &
  \begin{tabular}[c]{@{}c@{}}\textbf{GFLOP Per Second}\\ $1 \times 10^9$\end{tabular} \\ \hline
iMoT       & 14.49         & 7.79          \\
CTIN       & 0.56          & 7.27          \\
RoLSTM     & \textbf{0.21} & \textbf{7.17} \\
RoTCN      & 2.03          & 33.17         \\
RoResnet18 & 4.64          & 9.16          \\ \hline
\end{tabular}%
}
\end{table}

\end{document}